# Variational MCMC

**Nando de Freitas**    **Pedro Højen-Sørensen**[†]

UC Berkeley Computer Science
387 Soda Hall, Berkeley
CA 94720-1776 USA
{jfgf,jordan,russell}@cs.berkeley.edu

**Michael I. Jordan**    **Stuart Russell**

[†] Department of Mathematical Modelling
Technical University of Denmark
DK-2800 Kongens Lyngby, Denmark
phs@imm.dtu.dk

## Abstract

We propose a new class of learning algorithms that combines variational approximation and Markov chain Monte Carlo (MCMC) simulation. Naive algorithms that use the variational approximation as proposal distribution can perform poorly because this approximation tends to underestimate the true variance and other features of the data. We solve this problem by introducing more sophisticated MCMC algorithms. One of these algorithms is a mixture of two MCMC kernels: a random walk Metropolis kernel and a block Metropolis-Hastings (MH) kernel with a variational approximation as proposal distribution. The MH kernel allows one to locate regions of high probability efficiently. The Metropolis kernel allows us to explore the vicinity of these regions. This algorithm outperforms variational approximations because it yields slightly better estimates of the mean and considerably better estimates of higher moments, such as covariances. It also outperforms standard MCMC algorithms because it locates the regions of high probability quickly, thus speeding up convergence. We also present an adaptive MCMC algorithm that iterates between improving the variational approximation and improving the MCMC approximation. We demonstrate the algorithms on the problem of Bayesian parameter estimation for logistic (sigmoid) belief networks.

## 1 INTRODUCTION

Many problems arising in machine learning and decision theory can be interpreted as high-dimensional integration tasks. Bayesian computation would be trivial if we could calculate integrals appearing in the form of expectations, marginalization and normalization. For instance, the posterior distribution, which is the solution to the Bayesian learning problem, is obtained by computing the normalising integral. Similarly, in statistical mechanics, the Gibbs distribution is obtained by computing the partition function (the large normalising sum). The most popular integration methods for calculating these important target distributions are the Laplace method, variational approximation and MCMC simulation.

The Laplace method is an approximate integration method based on truncated Taylor expansions of the integrand. That is, by approximating the integrand with a tractable function, it becomes possible to evaluate the integral analytically. Unfortunately, in high dimensions, this will require the expensive computation of many cross-derivative terms. Moreover, it will provide poor approximation results unless the integrand is approximately log-quadratic.

Variational approximation also relies on approximating the integrand. Yet, it approximates the integrand by a lower bound that makes the integral tractable and that results in a lower bound on the integral. The approximation error is then minimized by maximising the lower bound. In other words, we replace the integration problem by an easier optimization problem. Variational methods have been shown to provide fast and reasonable approximate estimates in many scenarios (Jaakkola and Jordan 1999, Jordan, Ghahramani, Jaakkola and Saul 1999). However, variational approximations often result in algorithms that yield poor estimates of high order moments, such a covariances and kurtosis.

MCMC simulation is a powerful and accurate integration method (Gilks, Richardson and Spiegelhalter 1996, Robert and Casella 1999). Here, one draws a set of samples from the target distribution. This distribution is then approximated by an empirical distribution, whose support is the set of samples and whose range is the number of times each sample appears. Hence, the complex integrals are replaced by simple discrete sums. The big disadvantage of these methods is that it is often impossible to draw samples from the target distribution directly. This problem is circumvented by drawing samples from a proposal distribution, and then using Markov chain weighing mechanisms to ensure that these samples correspond to samples from the target distribution. The main difficulty with this approach is that the design of the proposal distributions is far from trivial. It varies with application domain and, if not done properly, the algorithms can take very long to converge (i.e., mix poorly).

To attack the problems inherent to variational and MCMC approximation simultaneously, we introduce a new class of MCMC algorithms that use variational approximations as proposal distributions. We show that naive algorithms exploiting this property can mix poorly, but address this problem by introducing more sophisticated MCMC kernels based on block sampling and mixtures of MCMC kernels. In particular, we use mixtures with variational kernels that allow the algorithm to detect the regions of high probability quickly and Metropolis kernels that enable it to explore the neighborhood of these regions. The resulting algorithm converges quickly to the regions of high probability and also yields reasonable approximations to the entire distribution of interest. Our approach makes it possible to combine variational and MCMC algorithms within a rigorous probabilistic setting so as to exploit the benefits of both approaches. We also introduce an adaptive variational MCMC scheme, whereby the MCMC simulation is used to improve the variational approximation, which in turn is used as proposal distribution. That is, each algorithm assists the other adaptively.

Recently, Ghahramani and Beal (2000) showed that using a variational approximation for mixtures of factor analyzers as the proposal for an importance sampler could lead to an improvement in the accuracy of the results. The approach we take here is more general and surmounts many of the problems encountered in the importance sampling approach.

We demonstrate the approach on the task of Bayesian parameter estimation of logistic (sigmoidal) belief networks with latent variables. This problem is of interest for several reasons. First, it exhibits nonlinearity and non-Gaussianity. Second, it includes the problems of logistic regression and classification with missing observations as a sub-case. Third, the noise is very uninformative and consequently one has to be very careful when applying model testing techniques such as cross-validation. Fourth, the model with hidden nodes is unidentifiable. Lastly, this type of network has important connections with research carried out in the area of neural computation.

## 2   VARIATIONAL METHOD

The aim of variational methods is to convert a complex problem into a simpler, more tractable problem; see for example (Jordan et al. 1999). The simpler problem is generally characterized by a decoupling of the degrees of freedom in the original problem. This decoupling is achieved by introducing an extra set of parameters, the so-called variational parameters. The variational parameters are then optimized so that the solution to the simpler problem resembles the solution to the complex problem. Convexity bounds play an important role in the variational paradigm. For example, in many models with hidden variables, the likelihood $p(\mathbf{x}^v|\boldsymbol{\theta})$ of the observed data $\mathbf{x}^v$ given the model parameters $\boldsymbol{\theta}$ cannot be easily evaluated because it requires the integration of the hidden variables $\mathbf{x}^h$. However, if we know a lower bound on the likelihood, we can maximize this bound to obtain an approximate solution. Lower bounds on the likelihood can be easily obtained using Jensen's inequality

$$\begin{aligned}
\log p(\mathbf{x}^v|\boldsymbol{\theta}) &= \log \mathbb{E}_{q(\mathbf{x}^h)}\left[\frac{p(\mathbf{x}|\boldsymbol{\theta})}{q(\mathbf{x}^h)}\right] \\
&\geq \mathbb{E}_{q(\mathbf{x}^h)}[\log p(\mathbf{x}|\boldsymbol{\theta})] - \mathbb{E}_{q(\mathbf{x}^h)}[\log q(\mathbf{x}^h)] \quad (1)
\end{aligned}$$

where $q(\mathbf{x}^h)$ is an arbitrary density over the hidden states with respect to the Lebesgue or counting measure. The right hand side is the negative Kullback Leibler divergence between $q$ and $p$ (that is, $-KL(q\|p)$) while the the last term is known as the entropy, $\mathcal{H}(q(\mathbf{x}^h)) \triangleq -\mathbb{E}_{q(\mathbf{x}^h)}[\log q(\mathbf{x}^h)]$, of the distribution $q$. It is clear, therefore, that maximising the lower bound is equivalent to minimising the Kullback Leibler divergence.

We choose a parametric form, $\widehat{q}(\mathbf{x}^h|\boldsymbol{\lambda})$, of $q(\mathbf{x}^h)$ that makes the right hand side of equation (1) easy to evaluate. The variational parameters $\boldsymbol{\lambda}$ can then be optimized to get a bound that is as tight as possible.

It may be impossible, in general, to choose a specific functional form of $\widehat{q}(\mathbf{x}^h|\boldsymbol{\lambda})$ that makes the evaluation of $\mathbb{E}_{\widehat{q}(\mathbf{x}^h|\boldsymbol{\lambda})}[\log p(\mathbf{x}|\boldsymbol{\theta})]$ tractable. However, additional flexibility can be introduced by lower bounding $p(\mathbf{x}|\boldsymbol{\theta})$ with a well-chosen function $\widehat{p}(\mathbf{x}|\boldsymbol{\theta},\boldsymbol{\xi})$, where $\boldsymbol{\xi}$ denotes an additional set of variational parameters. To summarize, the variational approach involves the following two steps:

1. Introduce the variational parameters $\boldsymbol{\xi}$ to make the conditional joint distribution of the hidden and visible variables, $p(\mathbf{x}|\boldsymbol{\theta})$, tractable.

2. Introduce the variational distribution $q$ with parameters $\lambda$ to make the conditional marginal distribution of the visible variables, $p(\mathbf{x}^v|\boldsymbol{\theta})$, tractable.

Following these steps, we can easily obtain unnormalized lower bounds on the likelihood distribution ($\widehat{p}(\mathbf{x}^v|\boldsymbol{\theta},\lambda,\boldsymbol{\xi}) \leq p(\mathbf{x}^v|\boldsymbol{\theta})$), posterior distribution ($\widehat{p}(d\boldsymbol{\theta}|\mathbf{x}^v,\lambda,\boldsymbol{\xi}) \leq p(d\boldsymbol{\theta}|\mathbf{x}^v)$) and marginal likelihood ($\widehat{p}(\mathbf{x}^v|\lambda,\boldsymbol{\xi}) \leq p(\mathbf{x}^v)$). The parameters $\boldsymbol{\theta}$, $\boldsymbol{\xi}$ and $\lambda$ can be computed by maximising the lower bound on the marginal likelihood. This computation can be carried out using an EM algorithm.

## 2.1 Logistic Belief Networks

For demonstration purposes, consider the logistic belief network shown in Figure 1. This network represents the factorized distribution

$$p(\mathbf{x}_{1:n_x}|\boldsymbol{\theta}) = \prod_{i=1}^{n_x} p(\mathbf{x}_i|\mathbf{x}_{\pi(i)},\boldsymbol{\theta}_i),$$

where $\mathbf{x}_{1:n_x} \triangleq \{\mathbf{x}_1,\mathbf{x}_2,\ldots,\mathbf{x}_{n_x}\}$ represents a stacked set of nodes, $\mathbf{x}_i$ denotes the variable associated with node $i$, $\mathbf{x}_{\pi(i)}$ denotes the parent nodes of node $i$, and $\boldsymbol{\theta}_i$ are some unknown parameters associated with node $i$. We partition the countable set of random variables $\mathbf{x}_i \in \mathcal{X}$ into a visible part, $\mathbf{x}_i^v \in \mathcal{X}^v$, and a hidden part, $\mathbf{x}_i^h \in \mathcal{X}^h$, such that $\mathcal{X} = \{\mathcal{X}^v \cup \mathcal{X}^h\}$. We assume that we have $T$ sets of measurements for the observed variables; that is $\mathbf{x}^v \triangleq \mathbf{x}_{1:n_{xv},1:T}^v \in (\mathcal{X}^v)^{n_{xv} \times T}$. The distribution of the random variable $\mathbf{x}_i$ is parameterized by $\boldsymbol{\theta}_i \in \mathbb{R}^{n_{\pi(i)}}$, where $n_{\pi(i)}$ is the number of variables on which $\mathbf{x}_i$ depends; that is the number of parent nodes in the case of a belief network. In general, the cardinality of $\boldsymbol{\theta}$ is $n_\theta \triangleq \sum_i n_{\pi(i)}$. We restrict the parameterization of the conditional probability distributions to the following Bernoulli family with a logistic mapping

$$p(\mathbf{x}_i|\mathbf{x}_{\pi(i)},\boldsymbol{\theta}_i) = \prod_{t=1}^{T} g(\varphi_{i,t}) = \prod_{t=1}^{T} \frac{1}{1+\exp(-\varphi_{i,t})},$$

where $\varphi_{i,t} \triangleq \mathbf{x}_{i,t}(\alpha + \boldsymbol{\theta}_i' \mathbf{x}_{\pi(i),t})$, $\mathbf{x}_i \in \{-1,1\}^T$ and $\alpha$ is assumed to be fixed. (Note that we only make the latter assumption for presentation purposes. One could always introduce an extra node fixed to 1 and treat $\alpha$ as an extra parameter.) To complete the specification of the Bayesian model, we assume a Gaussian prior $\mathcal{N}(\boldsymbol{\mu}_0,\boldsymbol{\Sigma}_0)$ on the parameters $\boldsymbol{\theta}_i$ and prior independence, that is $p(d\boldsymbol{\theta}) = \prod_i p(d\boldsymbol{\theta}_i)$.

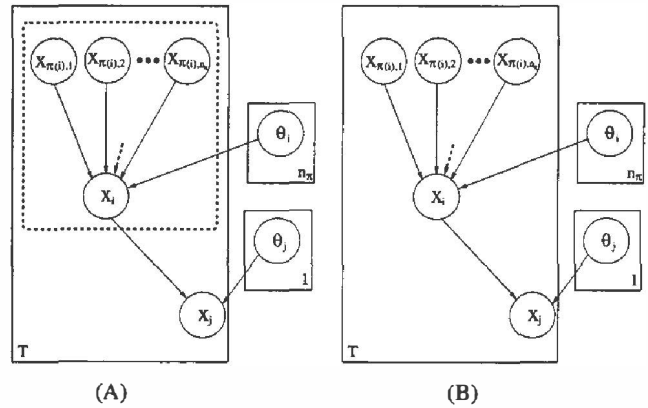

Figure 1: (A) Instance of a fully observed belief network. (B) Instance of a belief network with one hidden node (right). The parameters $\boldsymbol{\theta}$ are treated as hidden units in the Bayesian framework. The dashed box represents the Markov blanket for node $\boldsymbol{\theta}_i$, while the continuous boxes are templates indicating the number of copies of the variables inside them.

Note that the analysis applies to logistic BNs of arbitrary sizes and topologies. However, this problem often decouples into smaller sub-problems. When dealing with fully observed BNs, the Markov blanket of each $\boldsymbol{\theta}$ only depends on the data (see Figure 1) and, hence, we only need to solve several input-output logistic problems. This stops being the case when we have hidden variables. As shown in the example of Figure 1, if $\mathbf{x}_i$ is unobserved, nodes $\boldsymbol{\theta}_i$ and $\boldsymbol{\theta}_j$ become dependent. Yet, typically, one can still benefit from the structure in the network. The main difficulty arises when a node has many parents. In this paper, we focus on solving this problem.

The goal of the analysis will be to compute the posterior distribution $p(d\boldsymbol{\theta}|\mathbf{x}^v)$. From this distribution, one can easily derive other quantities of interest, such as predictive distributions and marginal distributions.

Following the variational procedure described at the beginning of this section, we place a tractable Gaussian lower bound on the likelihood (Jaakkola and Jordan 2000). This results on a Gaussian lower bound on the posterior ($\boldsymbol{\theta} \sim \mathcal{N}(\boldsymbol{\mu},\boldsymbol{\Sigma})$). We also assume that the hidden variables are decoupled and generated by a Bernoulli distribution ($\mathbf{x}^h|\lambda \sim \mathcal{B}e(\lambda)$). The EM equations that maximize the lower bound on the marginal likelihood, with $\phi(\boldsymbol{\xi}_i) \triangleq \frac{\tanh(\boldsymbol{\xi}_i/2)}{4\boldsymbol{\xi}_i}$, are

$$\boldsymbol{\Sigma}_{i,t}^{-1} = \boldsymbol{\Sigma}_{i,t-1}^{-1} + 2\phi(\boldsymbol{\xi}_{i,t-1})\mathbb{E}_{\widehat{q}(\mathbf{x}_i^h|\lambda_i)}\left[\mathbf{x}_{\pi(i),t}\mathbf{x}_{\pi(i),t}'\right]$$

$$\boldsymbol{\mu}_{i,t} = \boldsymbol{\Sigma}_{i,t}\left(\mathbb{E}_{\widehat{q}(\mathbf{x}_i^h|\lambda_i)}\left[\left(\frac{\mathbf{x}_{i,t}}{2} - 2\phi(\boldsymbol{\xi}_{i,t-1})\alpha\right)\mathbf{x}_{\pi(i),t}\right] + \boldsymbol{\Sigma}_{i,t-1}^{-1}\boldsymbol{\mu}_{i,t-1}\right)$$

$$\xi_{i,t}^2 = \alpha^2 + 2\alpha\mu_{i,t}'\mathbb{E}_{\widehat{q}(\mathbf{x}_i^h|\lambda_i)}\left[\mathbf{x}_{\pi(i),t}\right]$$
$$+ \text{tr}\left((\Sigma_{i,t} + \mu_{i,t}\mu_{i,t}')\mathbb{E}_{\widehat{q}(\mathbf{x}_i^h|\lambda_i)}\left[\mathbf{x}_{\pi(i),t}\mathbf{x}_{\pi(i),t}'\right]\right)$$
$$\lambda_j \approx \frac{\exp(D_j)}{1+\exp(D_j)}$$
$$D_j = \frac{\partial}{\partial\lambda_j}\mathbb{E}_{\widehat{q}(\mathbf{x}_{\pi(j)}|\lambda_j)}\left[\frac{\mathbf{x}_j\mu_j'\mathbf{x}_{\pi(j)} - \xi_i}{2}\right].$$

## 3　VARIATIONAL MCMC

The idea of Monte Carlo integration methods is to draw an i.i.d. set of samples $\{\theta^{(i)}; i = 1, 2, \ldots, N\}$ from the target distribution $p(d\theta)$ (it could be the posterior, $p(d\theta|\mathbf{x})$, in Bayesian analysis) to obtain the following empirical distribution

$$P_N(d\theta) = \frac{1}{N}\sum_{i=1}^{N}\delta_{\theta^{(i)}}(d\theta),$$

where $\delta_{\theta^{(i)}}(d\theta)$ denotes the delta-Dirac mass located in $\theta^{(i)}$. Consequently, one can approximate the integrals, $I(f)$, by discrete sums, $I_N(f)$, as follows

$$I_N(f) = \frac{1}{N}\sum_{i=1}^{N}f(\theta^{(i)}) \xrightarrow[N\to\infty]{a.s.} I(f) = \int_\Theta f(\theta)p(d\theta).$$

The estimate $I_N(f)$ is unbiased and by the strong law of large numbers, it will almost surely converge to $I(f)$. The main disadvantage of simple Monte Carlo methods is that often it is not possible to draw samples from $p(d\theta)$ directly. This problem can, however, be circumvented by the introduction of MCMC algorithms. Assuming that we can draw samples from a proposal distribution $\pi(d\theta)$, the key idea of MCMC simulation is to design Markov chain mechanisms that cause the proposed samples to migrate so that their empirical distribution approximates $p(d\theta)$.

The most popular example of this class of algorithms is the Metropolis-Hastings (MH) algorithm (Robert and Casella 1999). A Metropolis-Hastings step of invariant distribution, say $p(d\theta)$, and proposal distribution, say $\pi(d\theta^\star|\theta)$, involves sampling a candidate value $\theta^\star$ given the current value $\theta$ according to $\pi(d\theta^\star|\theta)$. The Markov chain then moves towards $\theta^\star$ with acceptance probability $\mathcal{A}(\theta,\theta^\star) = \min\{1, [p(d\theta)\pi(d\theta^\star|\theta)]^{-1}p(d\theta^\star)\pi(d\theta|\theta^\star)\}$, otherwise it remains at $\theta$. *It is well known that the success or failure of the algorithm often hinges on the choice of proposal distribution* (Gilks et al. 1996, Robert and Casella 1999).

The most obvious and immediate way of proceeding would be to sample new candidates according to the variational distribution. That is,

$$\pi(d\theta^\star|\theta) = \widehat{p}(d\theta^\star|\mathbf{x}^v, \mathbf{x}_\pi^v, \lambda, \xi).$$

In this case, the acceptance probability of the MH algorithm simplifies to $\mathcal{A}(\theta,\theta^\star) = \min\left\{1, \frac{w(\theta^\star)}{w(\theta)}\right\}$, where $w(\cdot) \triangleq p(\cdot)/\widehat{p}(\cdot)$ denotes the importance weights. This type of algorithm is known as the independent MH algorithm and it is closely related to the standard importance sampler (Geweke 1989). Both the importance sampler and independent MH algorithm are well known to perform poorly in high dimensions unless the proposal distribution is very close to the target distribution (Geweke 1989, Mengersen and Tweedie 1996). In particular, they work poorly when the proposal distribution underestimates the higher order moments of the target distribution. Unfortunately, this is one of the characteristics of the variational approximation. To address this problem, we introduce more sophisticated algorithms in the following subsections.

### 3.1　Block MCMC

To obtain a sampler that mixes faster, we can exploit the nature of the variational approximation and propose to update the parameters in blocks. Each proposal distribution corresponds to a Gaussian distribution whose mean is a subset of the elements of the mean of the original variational distribution and whose covariance is the corresponding block-diagonal component of the original covariance. The transition kernel, at iteration $(i + 1)$, for this algorithm is given by

$$K(\theta^{(i)}, A) = \prod_{j=1}^{n_b} K_{MH-j}(\theta^{(i)}_{b_{j-1}+1:b_j}, \theta^{(i+1)}_{-[b_{j-1}+1:b_j]}; A_j)$$

where $b_j$ denotes the size of the $j$-th block, $n_b$ denotes the number of blocks, $\theta^{(i+1)}_{-[b_j+1:b_{j+1}]} \triangleq \{\theta^{(i+1)}_{1:b_1}, \theta^{(i+1)}_{b_1+1:b_2}, \ldots, \theta^{(i+1)}_{b_{j-2}+1:b_{j-1}}, \theta^{(i)}_{b_j+1:b_{j+1}}, \ldots, \theta^{(i)}_{b_{n_b-1}+1:b_{n_b}}\}$ and $K_{MH-j}(\cdot; d\cdot)$ denotes the $j$-th MH algorithm in the cycle. (The Gibbs sampler is a special case of this scheme.) Since this kernel allows one to visit all sets of positive measure, while being aperiodic, simple convergence holds true as the number of samples becomes large. Obviously, choosing the size of the blocks poses some trade-offs. If one samples the components of a multi-dimensional vector one at a time, the chain may take a very long time to explore the target distribution. This problem gets worse as the correlation between the components increases. Alternatively, if one samples all the components together, then the probability of accepting this large move tends to be very low.

### 3.2　Mixtures of MCMC Kernels

A very powerful property of MCMC is that it is possible to combine several samplers into mixtures and cycles of the individual samplers (Robert and Casella 1999). This way we can have global proposals to explore large regions of the parameter space and local proposals to discover finer details of the target distribution (Andrieu and Doucet 1999, Andrieu, de Freitas and Doucet 2000). If the transition kernels $K_1$ and $K_2$ have invariant distribution $p(\cdot)$ each, then the *cycle hybrid kernel* $K_1K_2$ and the *mixture hybrid kernel* $\nu K_1 + (1-\nu)K_2$, for $0 \leq \nu \leq 1$, are also transition kernels with invariant distribution $p(\cdot)$. In this paper, we adopt a mixture where, with probability $\nu$, we sample $\theta$ using the variational block MCMC algorithm and, with probability $1 - \nu$, we carry out a random walk Metropolis step (also in blocks). The variational proposal locks into a region of high probability while the random walk allows one to explore the space around this region. This allows us to accomplish both rapid mixing and reasonable exploration of the target distribution.

### 3.3 Adaptive Variational MCMC

The goal of adaptation is to update the proposal distribution based on the behaviour of the Markov chain. That is, we start an MCMC simulation with an initial variational approximation, and then use the MCMC samples to update the variational approximation. This results on a better variational approximation and faster mixing. However, one should not allow adaptation to take place infinitely often as this can disturb the stationary distribution and the consistency results. This problem arises because by using the past information infinitely often, we violate the Markov property of the transition kernel. That is, $\Pr(\theta^{(i)}|\theta^{(0)}, \theta^{(1)}, \ldots, \theta^{(i-1)})$ does no longer simplify to $\Pr(\theta^{(i)}|\theta^{(i-1)})$. We can avoid this problem by performing adaptation only when the chain visits a particular atomic set. At this atomic set, the chain regenerates and, hence, the next tour becomes independent of the past tour. We adopt an algorithm based on this principle which was proposed by Gilks, Roberts and Sahu (1998).

## 4 SIMULATIONS

We performed experiments on fully and partially observed logistic BNs. When all the nodes are observed, the posterior is unimodal. This allows us to compare the algorithms in high dimensions by evaluating the distance between their estimates of the mean and the optimal mean. The likelihood, using a flat prior so that it is close to the posterior, will be higher for estimates close to the optimal posterior mean. We used this performance test because the optimal mean can be very different from the generating mean as shown in Figure 2. To illustrate this, we generated data 4 times using the same model with the parameter set to 1. Each realization of the data gave us a different likelihood (and posterior). Hence, if we were to have a model that represents the posterior well, it is not guaranteed to predict well. The noise model is too uninformative and, therefore, poorly suited to predictive testing techniques such as cross-validation. The problem is exacerbated as the dimension of the parameter space increases. We also performed experiments on multimodal distributions that show the performance of the algorithm not only in terms of approximating the mean, but in terms of approximating more complex aspects of the posterior distribution.

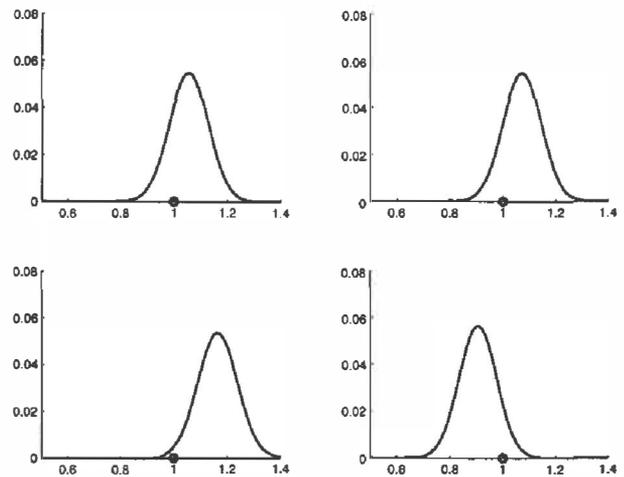

Figure 2: Likelihood of the data (1000 observations) when generated by a Bernoulli logistic node with a single parameter set to 1. Clearly, 1000 observations are not enough to recover the true value of the parameter. We are dealing with a very uninformative noise model.

### 4.1 Unimodal Models

We used logistic models consisting of one child and a varying number of parents, ranging from 1 to 50. We generated sets of 1000 data samples from these models. We computed posterior approximations using the variational EM algorithm, the block M-H sampler with the variational proposal distribution (VarMCMC), the random walk Metropolis (RW), and the MCMC mixture with a variational kernel and a Metropolis kernel (VarMixMCMC). We repeated this experiment 10 times to obtain estimates of the performance in terms of means and error bars. We set the random walk variance to 0.01, the bias parameter to 0.5, the Bernoulli mean to 0.5 and the generating parameters to uniformly random values on $(0, 1]$. We chose a fairly flat prior $\mathcal{N}(0, 100\mathbf{I})$. The results for 500 and 5000 samples are shown in Figures 3 and 4 respectively.

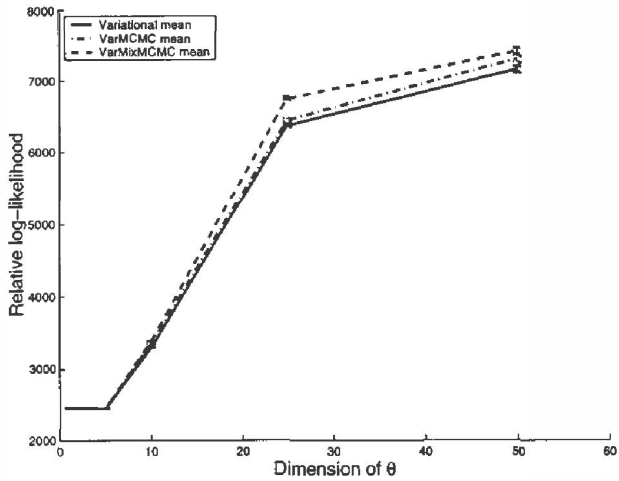

Figure 3: This figure shows the relative log-likelihood of the three variational methods with respect to the log-likelihood of the random walk metropolis (500 samples). Since all the curves are positive, the three methods outperform the metropolis algorithm. In addition, the MCMC mixture with variational and Metropolis kernels provides better estimates of the mean for different numbers of parents.

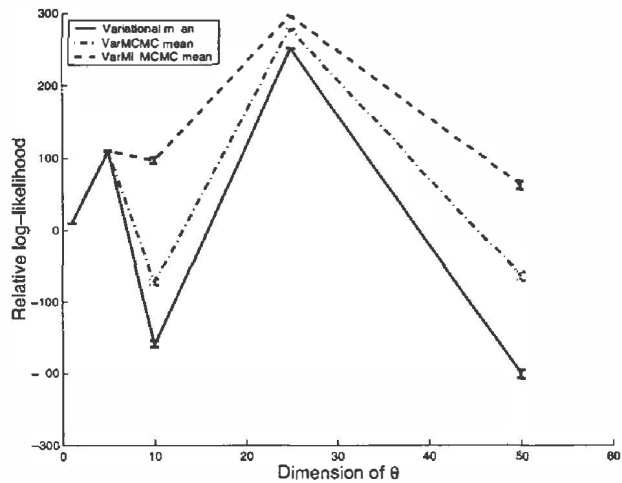

Figure 4: The MCMC mixture with variational and Metropolis kernels provides better estimates of the mean for 5000 samples. Note that although the performance of the Metropolis algorithm has improved it does not perform better than VarMixMCMC. Recall that VarMixMCMC has a random walk component and hence, at worst, will perform similarly to the standard Metropolis.

It is clear that the VarMixMCMC algorithm outperforms the VarMCMC algorithm, which in turn outperforms the standard variational algorithm. The performance of the RW depends on the initialization and data set realization. That is, it might or might not perform well depending on whether it is initialized in regions of high probability or not. Of course, as the number of samples goes to infinity, the RW algorithm will approximate the mean according to the central limit theorem. Yet, in practical scenarios we often need reliable and faster options. The computational time for the EM and MCMC algorithms is shown in Table 1.

|           | 500 samples ||  5000 samples ||
| Dimension | EM | MCMC | EM | MCMC |
| --- | --- | --- | --- | --- |
| 1  | 0.01 | 1.00 | 0.01 | 10.03 |
| 5  | 0.43 | 9.73 | 0.40 | 97.49 |
| 10 | 3.18 | 31.25 | 3.36 | 313.17 |
| 20 | 52.16 | 204.54 | 23.49 | 1237.87 |
| 50 | 477.34 | 1351.90 | 478.59 | 13081.37 |

Table 1: Computational time in Mega-flops for the EM (Variational) and MCMC algorithms.

In this experiment, we have discussed the performance of the methods only in terms of approximating the mean of the posterior. However, we often want to compute other characteristics of the distribution. In the following section, we show that the VarMixMCMC algorithm is well suited to this more difficult problem.

### 4.2 Multimodal Models

In this experiment, we considered a network with two parents (one hidden and one observed). The posterior for $\theta$ is, therefore, bivariate and can have two modes. (Note that the two modes appear because of unidentifiability. Either of these modes provides a statistically valid solution.) For demonstration, we set the generating parameters for the hidden and observed nodes to 2 and −1 and the respective Bernoulli means of the hidden variables to 0.6 and 0.5. We set the bias parameter to 2, the number of data 50 and the prior to $\mathcal{N}(3, 10\mathbf{I})$. The posterior in this case can be evaluated numerically on a two-dimensional grid. We show its contour curves in Figure 5. This figure also shows the contour plot of the RW MCMC histogram after 5000 iterations and the variational approximation. We notice that the variational approximation fits closely to one of the modes. We also notice that if the random walk starts in a region of low probability, it can take long to locate one of the modes. Its performance will, therefore, be poor when dealing with posteriors with elongated contours. Figure 6 illustrates the point that the naive variational MCMC algorithm locates one of the modes but fails to explore the support of the posterior. The mixture MCMC algorithm, shown in Figure 7 solves this problem and provides the best solution out of all the methods.

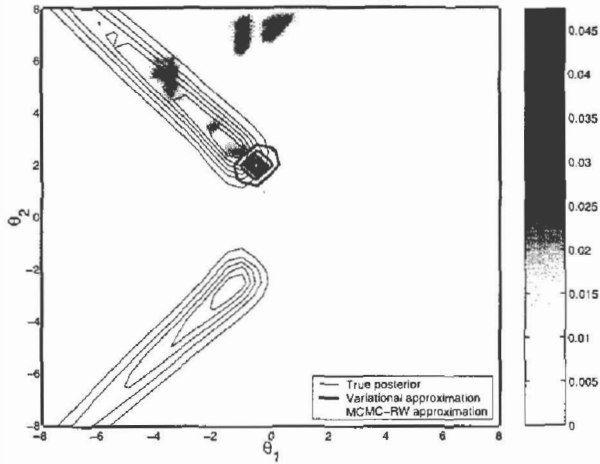

Figure 5: Approximation with the random walk Metropolis algorithm after 5000 iterations for a bivariate model. The contour plot of the 2D histogram of the MCMC samples, indicates that the random walk can spend a considerable time in regions of low probability.

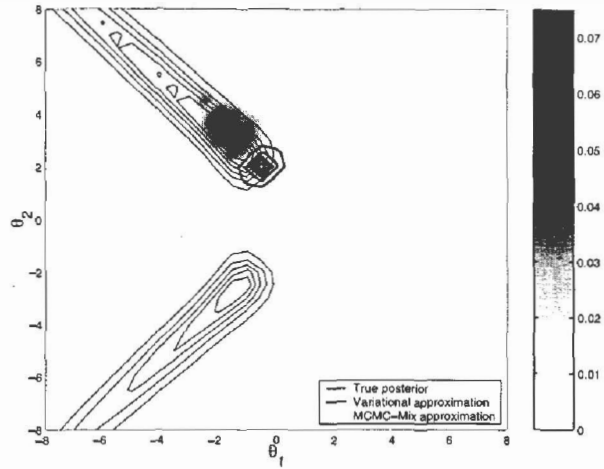

Figure 7: Approximation with the mixture MCMC algorithm after 5000 iterations for a bivariate model. The variational component allows us to locate a region of high probability and the random walk allows us to explore the neighborhood of this region.

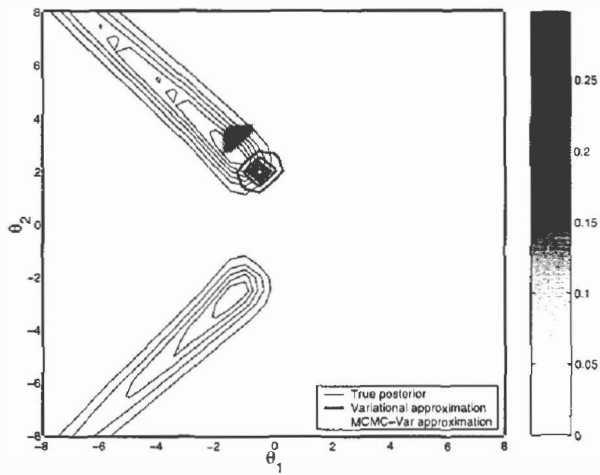

Figure 6: Approximation with the variational MCMC algorithm after 5000 iterations for a bivariate model. The variational approximation allows us to locate a region of high probability.

### 4.3 Adaptive MCMC Experiment

We tested our adaptive MCMC sampler through regeneration on the unimodal scenario. We found, as shown in Figure 8, that the algorithm works well; we end up with a variational approximation that does not underestimate the variance. Note, however, this method is appropriate only for problems with lim-ited fan-in—we found that the acceptance rate decays rapidly to zero beyond a fan-in of seven.

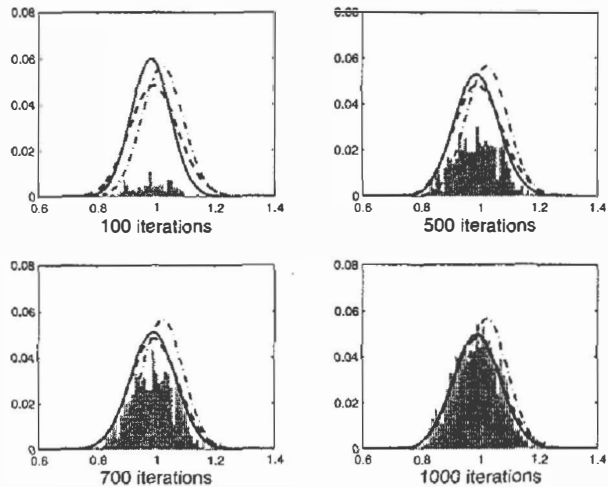

Figure 8: Adaptive MCMC. The variational approximation [-·] underestimates the variance of the true posterior [--]. It also exhibits a slightly different mean. Using the samples generated by the Markov chain we update the variational proposal. After only 100 iterations, the new variational approximation [-] already provides a better estimate of the mean. Eventually, the variational approximation becomes much closer to the target distribution and the MCMC algorithm converges well.

## 5 CONCLUSIONS AND FURTHER WORK

This paper demonstrates that it is advantageous to combine variational and MCMC methods. Variational methods allow us to map the problem under consideration to a subset of simpler problems. By solving these subproblems we obtain suboptimal distributions, that can in turn be used as proposals for more complex sampling schemes. We pointed out that naive algorithms based on this principle can perform poorly because the variational approximation tends to underestimate the variance of the posterior distribution. We therefore proposed more sophisticated MCMC algorithms that are clearly able to benefit from the variational approximation and outperform standard Metropolis algorithms.

In the multimodal scenario we focused on the problem of approximating only one of the modes. For many models, multimodality arises as the result of label permutation (unidentifiability) and hence any mode is a correct statistical solution. This is the case of mixture models. We do recognize that in more complex situations, where there are more sources of multimodality, we will need to extend our algorithms. One simple strategy is to compute several variational approximations using different initial conditions. These approximations can then be used either in parallel or in a multiple MCMC mixture to visit several modes quickly. The tempering method described in (Neal 1996) will also serve the purpose of jumping modes.

We feel that it is essential to carry out more research in the direction of adaptive MCMC. Ultimately we would like to represent high dimensional distributions with a mixture of adapted (better) variational approximations. In very large dimensional mixtures for document retrieval, one may require up to 100 megabytes to store a single sample (Hofmann and Puzicha 1998). The storage requirements would decrease considerably if we were able to only store the sufficient statistics. Needless to say, better proposal distributions will also lead to faster convergence and improved results.

There are a few more interesting research directions. First, we need to consider algorithms that exploit both lower and upper variational bounds. These, we believe, will allow us to locate modes and jump between them efficiently. Second, we only need to use the variational approximation to approximate one of the marginals. It is, therefore, possible to apply this idea when implementing complex hierarchical Bayesian schemes.

Lastly, a more detailed technical report is available at http://www.cs.berkeley.edu/~jfgf/publications.html.

### Acknowledgements

We would like to thank Christophe Andrieu, Mathew Beal, Arnaud Doucet, Zoubin Ghahramani, Tommi Jaakkola and Kevin Murphy.